\documentclass[journal]{IEEEtran}
\usepackage{amsmath,amsfonts}
\usepackage{algorithmic}
\usepackage{array}
\usepackage[caption=false,font=normalsize,labelfont=sf,textfont=sf]{subfig}
\usepackage{textcomp}
\usepackage{stfloats}
\usepackage{url}
\usepackage{verbatim}
\usepackage{graphicx}
\def\BibTeX{{\rm B\kern-.05em{\sc i\kern-.025em b}\kern-.08em
    T\kern-.1667em\lower.7ex\hbox{E}\kern-.125emX}}
\usepackage{balance}

\begin{document}

\title{Legal Evalutions and Challenges of Large Language Models}
\author{Jiaqi Wang$^*$, Huan Zhao$^*$\thanks{$^*$Equal contribution. Listing order is random.}, Zhenyuan Yang, Peng Shu, Junhao Chen, \
Haobo Sun, Ruixi Liang, Shixin Li, Pengcheng Shi, Longjun Ma, Zongjia Liu, Zhengliang Liu, Tianyang Zhong, Yutong Zhang, Chong Ma, Xin Zhang, Tuo Zhang, Tianli Ding, Yudan Ren, Tianming Liu$\dagger$, Xi Jiang$\dagger$, Shu Zhang$\dagger$\thanks{$\dagger$Corresponding authors: Tianming Liu, Xi Jiang, Shu Zhang.}
\thanks{Manuscript created October, 2020; This work was developed by the IEEE Publication Technology Department. This work is distributed under the \LaTeX \ Project Public License (LPPL) ( http://www.latex-project.org/ ) version 1.3. A copy of the LPPL, version 1.3, is included in the base \LaTeX \ documentation of all distributions of \LaTeX \ released 2003/12/01 or later. The opinions expressed here are entirely that of the author. No warranty is expressed or implied. User assumes all risk.}}

\markboth{Journal of \LaTeX\ Class Files,~Vol.~18, No.~9, September~2020}%
{How to Use the IEEEtran \LaTeX \ Templates}

\maketitle

\begin{abstract}
In this paper, we review legal testing methods based on Large Language Models (LLMs), using the OPENAI o1 model as a case study to evaluate the performance of large models in applying legal provisions. We compare current state-of-the-art LLMs, including open-source, closed-source, and legal-specific models trained specifically for the legal domain. Systematic tests are conducted on English and Chinese legal cases, and the results are analyzed in depth. Through systematic testing of legal cases from common law systems and China, this paper explores the strengths and weaknesses of LLMs in understanding and applying legal texts, reasoning through legal issues, and predicting judgments. The experimental results highlight both the potential and limitations of LLMs in legal applications, particularly in terms of challenges related to the interpretation of legal language and the accuracy of legal reasoning. Finally, the paper provides a comprehensive analysis of the advantages and disadvantages of various types of models, offering valuable insights and references for the future application of AI in the legal field.

\end{abstract}


\section{Introduction}

In recent years, the breakthrough of deep learning technology in natural language processing (NLP), particularly the rapid advancement of Transformer technology, has led to the flourishing of LLMs~\cite{wang2023review}. Models like OpenAI's GPT series have demonstrated exceptional capabilities in NLP, excelling not only in traditional NLP tasks such as machine translation and Question Answering, but also in some multimodal tasks, such as image-to-text translation, speech recognition, and subtitle generation~\cite{wang2024comprehensive,brown2020language,vaswani2017attention,radford2018improving}. These models are capable of accurately understanding relationships between various data forms and enabling cross-modal information transformation, significantly enhancing automation and efficiency across these fields.

In the legal field, LLMs are seen as a transformative force with the potential to revolutionize traditional legal services, owing to their comprehensive legal knowledge base and exceptional capabilities in natural language understanding and generation~\cite{lai2024large}. Some studies have explored the application of LLMs in the analysis and generation of legal texts, evaluating their performance in tasks such as legal reasoning, case retrieval, and legal question answering, and investigating their potential to improve the efficiency and accuracy of legal work~\cite{chaudhary2024judge}. Meanwhile, other researchers have focused on developing LLMs specifically tailored for legal domains, enabling these models to better understand legal terminology, apply legal provisions accurately, and adapt to the nuances of different legal systems. This specialization aims to increase the practical value of LLMs in legal practice~\cite{zhou2024lawgpt,chalkidis2023chatgpt,LAWGPT_zh}. However, effectively evaluating the performance of LLMs across various legal systems and linguistic environments remains a significant challenge. Additionally, addressing the technical and ethical concerns associated with their application is an urgent issue that requires further attention and resolution.

The application of LLMs in the legal field also faces numerous challenges and issues. First, legal language is highly specialized and precise, making it crucial to ensure the accuracy and legality of the content generated by these models~\cite{danet1980language,guha2024legalbench}. Second, LLMs may absorb biases and inaccuracies from their training data, which can have serious repercussions when applied in the legal context~\cite{hadi2024large,ray2023chatgpt,cheong2024safeguarding}. Additionally, the automation of legal decision-making processes could lead to ethical concerns and disputes over legal accountability~\cite{osasona2024reviewing,akpuokwe2024legal}.

\begin{figure*}[ht]
\begin{center}
\includegraphics[width=1.0\textwidth]{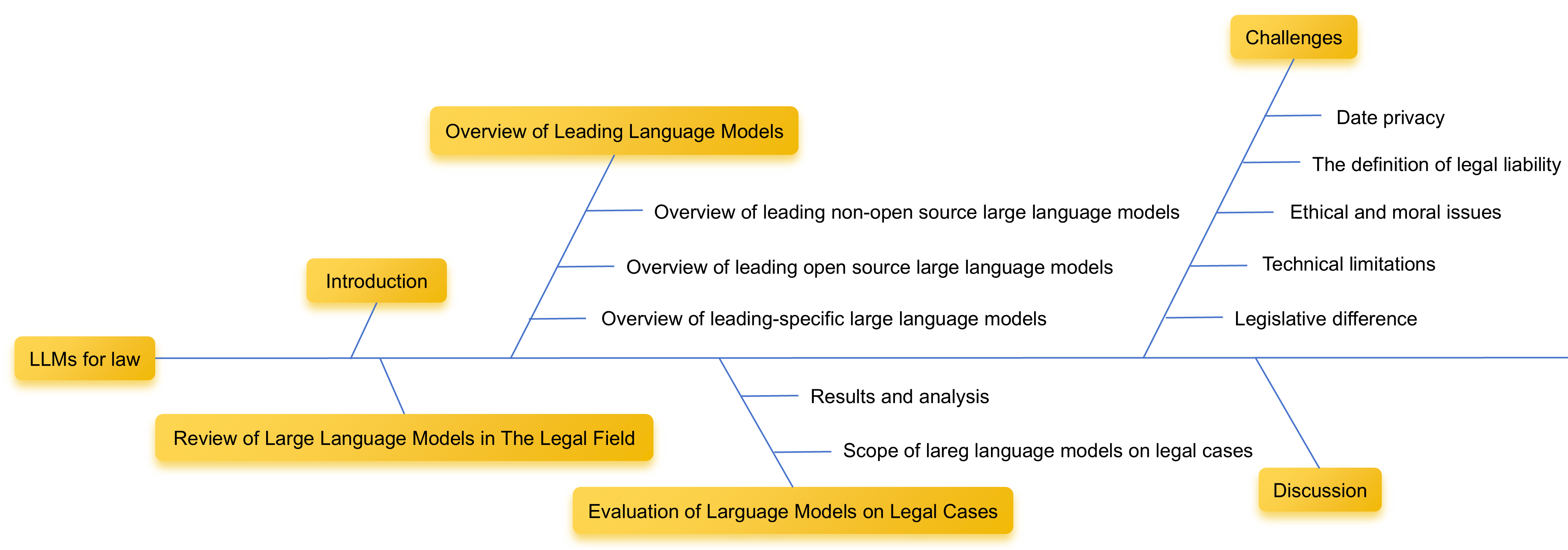}
\end{center}
\caption{Overview of the study
} 
\label{outline}
\end{figure*}

As shown in Fig~\ref{outline} Based on this background, this work aims to provide a comprehensive overview of the performance of LLMs in the legal field, offering valuable insights for both the academic community and legal practitioners. The study is structured as follows:

Section 1: This Section explains the background, purpose, and significance of the study, outlining the motivations and objectives behind the research.

Section 2: This Section provides a detailed analysis of legislation related to large models on a global scale, exploring the similarities and differences in policies and regulations across various countries.

Section 3: The focus is on models specifically tailored to the legal domain, examining their technical features and evaluating their potential applications in legal practice.

Section 4: This Section presents a comprehensive assessment of the models discussed in Section 3 using thirteen Chinese and thirteen English legal cases. The cases were selected to include a complete set of four components: judgment, background, analysis and conclusion. The Section systematically evaluates the performance and applicability of each model through a comparative analysis of results and quantitative metrics.

Section 5: This Section discusses key issues related to the use of LLMs in the legal field, including data privacy, legal liability, ethical considerations, and technical limitations.

Section 6: The final Section summarizes the findings of the study and provides an outlook on future research directions.

Through this study, we aim to provide in-depth insights into the application of LLMs in the legal area, fostering their rational and sustainable integration into legal practice. This will not only contribute to improving the efficiency and quality of legal services but also lay a solid foundation for future innovations in legal technology.

\section{Review of LLMs in the legal field}
The rapid advancement of LLMs has catalyzed significant breakthroughs in NLP and across sectors like medical healthcare \cite{pan2024eg,li2024echopulse,shi2024mgh}, education \cite{shu2024llms, tian2024assessing, lee2023multimodality}. Encouraged by these successes, researchers are increasingly exploring LLM applications in the legal domain. LLMs hold substantial potential to assist legal professionals in tasks such as summarization, drafting (e.g., contract clauses or initial document drafts), and legal research \cite{nicola2023}. Summarization tasks can range from generating concise contract summaries \cite{tom2023}, summarizing complex litigation filings in case dockets \cite{stephanie2023}, to producing automatic summaries of judicial opinions \cite{monde2023}. In drafting, LLMs can review and suggest language improvements in documents and contracts \cite{taher2024}, as well as enrich drafting options \cite{Henchmanr2024}—for example, modifying clauses to switch between singular and plural forms or appending additional elements \cite{Henchmanr2024}. Legal research applications leverage LLMs to provide plain language responses to legal queries, synthesizing case law and offering clear, accessible answers to intricate legal questions, including those related to securities law \cite{greg2023}. Furthermore, LLMs can generate tailored research memoranda in response to specific queries \cite{blue2023} and facilitate the development of chatbots capable of answering questions on Supreme Court rulings. These capabilities underscore LLMs' transformative role in enhancing efficiency, precision, and accessibility in legal practice.

Research has increasingly focused on assessing the capabilities of LLMs in the legal domain. For instance, \cite{chalkidis2023chatgpt} investigated the zero-shot performance of GPT-3.5 Turbo on the LexGLUE benchmark \cite{chalkidis2021lexglue}, utilizing a templated, instruction-based approach. Their findings indicate that while ChatGPT achieves an average micro-F1 score of 49.0\% across LexGLUE tasks—surpassing baseline guessing rates—it still demonstrates overall poor performance in legal text classification, suggesting limitations in handling nuanced legal language.

Extending these evaluations, \cite{deroy2023ready} examined the potential of LLMs for generating abstractive summaries of case judgements, applying both domain-specific and general-domain models to Indian court rulings. Their results indicate that while these models can generate coherent summaries, neither pre-trained abstractive summarization models nor general-purpose LLMs are yet suitable for fully automated case judgement summarization, due to quality inconsistencies and domain-specific limitations.

Similarly, \cite{savelka2023explaining} assessed GPT-4’s performance in generating precise, relevant explanations for legal terminology, specifically within legislation. Although initial impressions of GPT-4’s output were favorable, closer analysis revealed inaccuracies, highlighting current limitations in factual precision within legal text generation. Further research on structured improvements may thus be needed before deploying LLMs for critical, domain-specific tasks like case summarization and legislative interpretation.

To explore broader capabilities, \cite{simmons2023garbage} leveraged the commonsense reasoning abilities of LLMs for zero-shot crime detection based on descriptive summaries of surveillance videos. Their study underscores that, given accurate textual descriptions, LLMs achieve state-of-the-art results in crime detection and classification through zero-shot reasoning. However, they identify that the accuracy of video-to-text conversion remains a significant obstacle to practical deployment.

These aforementioned models have been applied extensively to legal tasks. However, their performance is often constrained when relying on zero-shot settings, which limits their ability to fully leverage domain-specific knowledge. To address this limitation, various efforts have focused on developing advanced legal LLMs by utilizing large-scale legal data for continuous pre-training and supervised fine-tuning.

For example, LAWGPT-zh is an open-source Chinese legal language model based on ChatGLM-6B and fine-tuned through 16-bit LoRA instruction. This model incorporates a substantial legal question-and-answer dataset, built from both legal articles and practical case studies, aimed at enhancing legal consultation capabilities \cite{LAWGPT_zh}. Similarly, LAWGPT \cite{zhou2024lawgpt} represents one of the first open-source models tailored for Chinese legal applications, which leverages large-scale Chinese legal documents for domain-specific pre-training. This approach enables the model to incorporate legal knowledge and improve its performance across various downstream tasks by creating a knowledge-driven, supervised fine-tuning dataset.

Other models such as Lawyer-LLama \cite{huang2023lawyer, Lawyer_LLama} have also emerged, with a focus on mastering Chinese legal knowledge and providing accessible explanations for legal concepts. This model spans areas such as marriage, lending, maritime, and criminal law, aiming to deliver essential legal consultation. LexiLaw \cite{LexiLaw2023} is similarly based on ChatGLM-6B architecture but is fine-tuned specifically to enhance legal consultation and support through targeted legal datasets.

LexGPT 0.1 \cite{lee2023lexgpt}, developed using GPT-J and pre-trained with Pile of Law, allows legal professionals to customize LLMs for downstream legal tasks with minimal technical requirements. Meanwhile, ChatLaw \cite{cui2023chatlaw} has been designed to reduce hallucination risks during legal data retrieval by combining vector database and keyword-based retrieval, thus improving the reliability of reference data.

DISC-LawLLM \cite{yue2023disc} takes a comprehensive approach by integrating legal syllogism-based reasoning to enhance its understanding of Chinese legal knowledge, while also incorporating a retrieval module to further support background knowledge adherence. KL3M \cite{KL3M2024}, the Kelvin Legal LLM, represents a pioneering "from-scratch" model for legal, regulatory, and financial applications in enterprise settings, built on clean, permissible data for enhanced utility and compliance.

In addition to these models, there are numerous other legal LLMs trained on extensive legal datasets \cite{fei2024internlm, colombo2024saullm, wisdomInterrogatory2024, deng_etal_2023_syllogistic}, each leveraging specialized pre-training and fine-tuning strategies to maximize their applicability and accuracy within legal contexts. These advancements illustrate the trajectory of LLM development in the legal domain, with increasing focus on domain-specific optimization, reduced hallucination, and reliable consultation capabilities.
\section{Overview of leading language models}
\subsection{Overview of leading non-open source LLMs}
In recent years, with the advancement of computing power and the accumulation of massive amounts of data, LLMs have demonstrated immense potential in the field of artificial intelligence. They have made significant strides in various domains such as natural language processing and computer vision, capable of handling complex tasks like text generation, image recognition, and machine translation. Closed-source models like OpenAI's GPT-4 \cite{achiam2023gpt}, with their massive parameter counts and high-quality training data, have showcased exceptional abilities in understanding and generating human language, setting new benchmarks for AI technology. However, their capabilities in legal case adjudication remain to be explored.

GPT-4\cite{achiam2023gpt} is the fourth iteration of the Generative Pre-trained Transformer (GPT) series developed by OpenAI. With a colossal 1.8 trillion parameters, it significantly surpasses its predecessors. GPT-4 employs 16 mixed-expert models, each consisting of 1.11 trillion parameters. Trained on massive amounts of multimodal data, GPT-4 exhibits exceptional performance in tasks such as text generation and image understanding. Notably, GPT-4 possesses emergent abilities, enabling it to learn complex patterns from data without explicit programming, leading to more flexible and powerful task handling. GPT-4o, an optimized version of GPT-4, builds upon its predecessor's strengths and introduces technical improvements to significantly enhance efficiency and cost-effectiveness in specific scenarios, making it more suitable for practical applications.

Gemini, a multimodal LLM developed by Google AI, demonstrates exceptional performance in processing text, images, and other modalities. By directly mixing different modalities during pre-training, Gemini \cite{team2023gemini} establishes a deep understanding of the relationships between them. Gemini 1.5 further enhances its capabilities by supporting ultra-long contexts of up to millions of tokens. To improve efficiency and scalability, Gemini 1.5 \cite{team2024gemini} leverages a Mixture-of-Experts (MoE) architecture and is trained on Google's TPU v5e chips. This design enables the model to handle complex tasks efficiently and accurately. Gemini represents a significant advancement in the field of AI, paving the way for new applications and possibilities.

Claude 3.5 Sonnet\cite{anthropic2024claude}, developed by Anthropic, is a powerful language model that strikes a balance between speed and performance. Positioned as an intermediate model in the Claude 3 series, it offers exceptional coding and visual processing capabilities while maintaining efficiency. With an ultra-long context window of 200K tokens, the model can handle complex and lengthy legal texts and has outperformed its peers in various benchmarks. Its unique ability to "control a computer" gives it a distinct advantage in legal case analysis, enabling it to interact with computers like a human and process cases involving multimodal information such as images and diagrams. Moreover, the model has been carefully designed with security in mind, meeting the confidentiality requirements of legal case analysis. The emergence of Claude 3.5 Sonnet opens up new possibilities in legal AI, promising to play a significant role in legal text analysis, contract review, and case law retrieval.

Yi-Large\cite{young2024yi} is a LLM designed to handle multimodal data, including text and images. It incorporates Vision Transformer (ViT) and text encoders to achieve deep fusion of visual and textual features, enabling the model to understand and reason about multimodal information. With a context window of 200K tokens, Yi-Large can process long sequences effectively. To improve efficiency and performance, Yi-Large adopts grouped query attention and a three-stage training strategy. Experimental results show that Yi-Large outperforms state-of-the-art models on various multimodal tasks, including visual question answering and image generation.

\subsection{Overview of leading open source LLMs}
While closed-source models like OpenAI's GPT-4 have demonstrated exceptional performance in the realm of LLMs, the contributions from the open-source community are equally noteworthy. Open-source LLMs, such as Meta's Llama 3\cite{dubey2024llama} and models from Mistral AI\cite{jiang2023mistral}, have provided researchers and developers with vast opportunities for innovation due to their openness and accessibility. These models excel in various tasks including text generation and translation, and in some cases, their performance is on par with closed-source models. In the legal domain, open-source models have also shown immense potential. By learning from massive amounts of legal text, these models can provide strong support for legal research and practice. To thoroughly evaluate the application prospects of these models in the legal field, we have conducted in-depth research on the current mainstream open-source models.

Meta's newly released Llama 3 LLM\cite{dubey2024llama} marks a significant advancement in the field of AI. Built upon the auto-regressive Transformer architecture, Llama 3 incorporates optimizations in tokenization, attention mechanisms, and other key components. Through techniques such as supervised fine-tuning and reinforcement learning from human feedback, Llama 3 has achieved notable improvements in both performance and safety. Capable of handling multiple languages, long-form text, and complex reasoning, Llama 3 excels in tasks ranging from mathematical problem-solving to legal text analysis. Its open-source nature fosters innovation by empowering developers to customize the model for specific applications. By demonstrating state-of-the-art performance across various benchmarks, Llama 3 solidifies Meta's position as a leader in AI research. Moreover, Meta's commitment to responsible AI development is exemplified by the safety measures integrated into Llama 3. With its potential to revolutionize fields such as legal research, contract analysis, and case law retrieval, Llama 3 represents a promising step forward in the evolution of LLMs.

Mistral AI, a burgeoning AI startup founded by former employees of DeepMind and Meta, has made significant strides in the field of LLMs. Within a year of its inception, Mistral AI unveiled its inaugural model, Mistral 7B\cite{jiang2023mistral}, which promptly outperformed all other open-source models of the same parameter scale. Remarkably, it even surpassed larger models, demonstrating superior performance in tasks such as reasoning, mathematics, and code generation. Subsequent iterations, including Mistral 8x7B and Mistral Large 240B, have continued to push the boundaries of LLM capabilities, closing the gap with industry benchmarks like GPT-4. These models leverage advanced techniques such as GQA, RoPE, and SWA to enhance their ability to process long texts, perform complex reasoning, and generate code. The rapid growth and exceptional performance of Mistral AI have garnered significant attention within the AI community. By pioneering innovative approaches to LLMs, Mistral AI is shaping the future of natural language processing.

Gemma\cite{team2024gemma} is an open-source family of models based on Google's Gemini model, inheriting its strong generalization, understanding, and reasoning abilities. Trained on a massive dataset of up to 6 trillion tokens, the Gemma family achieves remarkable results in text generation, understanding, and reasoning. The series offers two model sizes, 7 billion and 20 billion parameters, to cater to various computational resources and application scenarios. Gemma 2\cite{team2024gemma2}, the latest addition to the series, adopts a decoder-only architecture and introduces several innovative techniques such as sliding window attention, soft-max, RMSNorm normalization, and grouped query attention, further enhancing the model's performance and efficiency. These innovations enable Gemma 2 to handle longer context windows while maintaining powerful language capabilities and improving training stability. The open-source nature of the Gemma models provides researchers and developers with a powerful tool, driving advancements in natural language processing.

Microsoft's newly released open-source Phi-3.5 series\cite{abdin2024phi} of AI models have achieved significant breakthroughs in performance and functionality. Among them, Phi-3.5-mini-instruct, designed for resource-constrained environments, excels in code generation and mathematical reasoning. Phi-3.5-MoE-instruct adopts a Mixture-of-Experts (MoE) architecture, ensuring efficient computation while handling complex tasks. Phi-3.5-vision-instruct combines text and image processing capabilities, demonstrating superior performance on multi-modal tasks. This model series has surpassed competing products in multiple benchmarks, setting new performance standards.

Qwen2\cite{yang2024qwen2} is a family of LLMs encompassing a wide range of parameter sizes, from 0.5B to 72B. This series excels in multilingual support, handling extra-long contexts, and computational efficiency. Built upon the Transformer architecture, Qwen2 incorporates techniques such as SwiGLU activation, QKV bias, and a mixture of SWA and Full Attention to enhance performance. Supporting 29 languages including Chinese and English, the model can process up to 128K tokens. Additionally, all models in the Qwen2 series employ the Grouped Query Attention (GQA) mechanism to reduce computational complexity and improve efficiency. These features make Qwen2 highly suitable for natural language processing tasks that require multilingual support, long-text processing, and complex reasoning.

The GLM-4 series\cite{glm2024chatglm}, developed by Zhipu AI, is a state-of-the-art family of pre-trained language models, offering various parameter sizes to cater to diverse application needs. This series excels in multilingual support, extra-long context processing, and multi-modal capabilities. GLM-4-9B and its dialogue variant, GLM-4-9B-Chat, outperform their counterparts in semantics, mathematics, reasoning, coding, and knowledge. GLM-4-9B-Chat further offers advanced functionalities such as web browsing, code execution, and custom tool calling. To address the need for extremely long context processing, we have introduced GLM-4-9B-Chat-1M, which supports a context length of up to 1 million tokens. Additionally, GLM-4V-9B, the multi-modal variant, demonstrates superior performance in bilingual (Chinese and English) multi-turn dialogue and image understanding, surpassing competitors including GPT-4-turbo. The open-source nature of the GLM-4 series makes it highly promising for both academic and industrial applications.

\subsection{Overview of legal-specific LLMs}
The legal domain demands a high degree of specialization from its models. Beyond general-purpose LLMs, we have evaluated models specifically tailored for legal tasks. These models, fine-tuned on extensive legal corpora, exhibit superior capabilities in understanding legal concepts, conducting legal reasoning, and generating legal text. Evaluating these models not only helps us assess their potential applications in the legal field but also provides valuable insights for advancing the development of legal artificial intelligence.

LexNLP\cite{bommarito2021lexnlp} is an open-source natural language processing toolkit specifically designed for legal text. It offers a comprehensive suite of text analysis capabilities, including text cleaning, tokenization, feature extraction, entity recognition, and text classification, enabling deep understanding of complex legal terminology and structures. Its modular design and flexible API allow users to customize functionalities based on their specific needs and seamlessly integrate it into various legal applications. LexNLP's strength lies in its profound understanding of legal text and its efficient information extraction capabilities, making it a valuable tool for legal research, contract analysis, and regulatory compliance.

Designed as a versatile legal language model, LawGPT\cite{zhou2024lawgpt} is fine-tuned on ChatGLM-6B LoRA 16-bit instructions and trained on a substantial corpus of Chinese legal text. The model has been enhanced with ChatGPT to refine and expand its training data, enabling it to provide comprehensive and accurate responses to complex legal inquiries. Moreover, LawGPT is being developed with a specialized legal knowledge base and a reliable self-instruction method to ensure the highest quality of legal advice. Distinguished by its exceptional performance in the Chinese legal domain, LawGPT offers a deeper understanding of Chinese legal nuances and provides more precise legal recommendations compared to other models.

ChatLaw\cite{cui2023chatlaw} is a cutting-edge legal AI assistant that combines knowledge graphs, mixed expert models, and multi-agent systems to provide comprehensive legal services. The ChatLaw model family includes a diverse range of models, from the BERT-based ChatLaw-Text2Vec to the large-scale pre-trained models ChatLaw-13B and ChatLaw-33B. Through extensive training on high-quality legal datasets, ChatLaw has developed exceptional capabilities in addressing complex legal questions and conducting in-depth legal reasoning. The ChatLaw2-MOE model, in particular, leverages a mixture-of-experts approach and a multi-agent system to enhance accuracy and reliability, surpassing other models including GPT-4 in various legal benchmarks. ChatLaw is particularly well-suited for the Chinese legal landscape, offering users tailored and expert legal advice.

\section{Evaluation of LLMs on legal cases}

\subsection{Scope of the Study and Used Datasets}
This study selected 26 representative legal cases as research subjects, with 13 from China and 13 from the United States, respectively using Chinese and English. To ensure the objectivity and fairness of the research, we strictly anonymized all personal privacy information in the cases.

The Chinese case dataset was constructed based on the Chinese Judgments Online database, covering civil, criminal, and administrative cases, and including a variety of judicial documents such as first-instance judgments, second-instance judgments, and rulings. The establishment of this dataset aims to comprehensively present the application of laws, judicial standards, and standardized expressions of judicial documents in various types of cases in Chinese judicial practice. Each case in the dataset contains detailed information, including basic information such as case number, court, trial date, and party identity, as well as the background, disputed issues, court's interpretation of legal provisions, evidence review, and final judgment with reasons for the case. Through this dataset, we can gain a deep understanding of the operation of the Chinese judicial system and provide rich data support for the study of Chinese law.

The US case dataset is sourced from the well-known Court Listener legal database, which collects a large number of judgment documents from federal and state courts in the United States. We carefully selected 13 representative cases from this vast database, covering multiple important legal areas such as immigration, criminal law, and administrative law. Each case provides rich and detailed information, including the social and legal background of the case, the focal issues that have attracted public attention, the core legal issues disputed by both parties, the final judgment of the court and detailed reasons for the judgment. If the case involves an appeal, we will also describe in detail the appeal process and the judgment of the higher court. In addition, the judgment provides related laws, regulations, precedents, and scholarly opinions to enable readers to conduct more in-depth research and understanding of these cases. Through this dataset, we can comprehensively understand the practices and basis of the US judicial system in handling different types of cases.

By comparing and studying the legal cases of China and the United States, we can not only deeply examine the large model's understanding and application capabilities in different legal systems but also deeply understand the similarities and differences between the two countries in terms of legislative concepts, judicial practices, and legal culture, and analyze the convergence and divergence of different legal systems in facing common legal issues in a globalized context. This research provides valuable experience for the application of large models in the legal field and can also provide rich first-hand data for legal scholars, judges, lawyers, and others, thereby promoting the continuous improvement of legal theory research and judicial practice.
\subsection{Results and Analysis}

In this section, we evaluate the performance of various LLMs (LLMs) on legal case judgment tasks, using both algorithmic and human evaluation metrics. We tested state-of-the-art models, including open-source, closed-source, and legal domain-specific models, across Chinese and English legal texts. For each model, the performance is assessed using the following metrics:

\textbf{ROUGE and BLEU Scores:} Algorithmic metrics like ROUGE and BLEU scores, both ranging from 0 to 1, are commonly used to evaluate text similarity between generated and reference outputs. ROUGE measures the overlap of n-grams between the model's output and a reference legal text, while BLEU calculates a modified form of precision for the generated output in comparison with human-generated text. Higher scores indicate closer alignment with reference cases, suggesting better model accuracy in generating relevant legal content.

\textbf{Human Evaluation Score:} To supplement automated metrics, we conducted a human evaluation to assess the quality of the model-generated judgments against actual case judgments made by legal professionals. Law students, trained in legal analysis, scored each model's decision output on a scale from 1 to 5, with 5 representing a high degree of alignment with the legal reasoning and outcomes in real-world cases. Human scores offer insight into how well model outputs mimic human judgment in complex legal scenarios.

In the results tables,  scores are reported for each model across \textbf{Chinese}, \textbf{English}, and \textbf{All} cases, representing performance in Chinese and English texts as well as an overall average.

\subsubsection{Performance on Chinese Legal Texts}

We evaluated the performance of various LLMs on Chinese legal texts using the metrics ROUGE-1, ROUGE-2, ROUGE-L, BLEU, and human evaluation scores. The results are summarized in Table \ref{tab:chinese_performance}.

\begin{table*}[ht]
\centering
\caption{Performance of LLMs on Chinese Legal Texts}
\label{tab:chinese_performance}
\begin{tabular}{|l|c|c|c|c|c|}
\hline
\textbf{Model} & \textbf{Chinese\_ROUGE-1} & \textbf{Chinese\_ROUGE-2} & \textbf{Chinese\_ROUGE-L} & \textbf{Chinese\_BLEU} & \textbf{Chinese\_Evaluation} \\ \hline
Gemma2-9B & 0.39 & 0.15 & 0.39 & 0.03 & 3.00 \\ \hline
GLM-4-9B-chat & 0.29 & 0.16 & 0.24 & 0.00 & 3.15 \\ \hline
GPT-4o & 0.13 & 0.01 & 0.10 & 0.00 & 3.85 \\ \hline
LawGPT\_zh & 0.27 & 0.08 & 0.16 & 0.04 & 1.85 \\ \hline
lawyer-llama-13b-v2 & 0.32 & 0.19 & 0.32 & 0.05 & 2.92 \\ \hline
llama3.2-3B-instruct & 0.30 & 0.11 & 0.15 & 0.04 & 1.62 \\ \hline
Mistral-7B-instruct-v0.3 & 0.38 & 0.15 & 0.20 & 0.07 & 2.54 \\ \hline
O1-preview & 0.13 & 0.02 & 0.09 & 0.00 & 3.85 \\ \hline
Phi-3.5-mini-instruct & 0.38 & 0.13 & 0.38 & 0.03 & 2.15 \\ \hline
Qwen2-7B-Instruct & 0.27 & 0.16 & 0.23 & 0.00 & 3.85 \\ \hline
\end{tabular}
\end{table*}

\textbf{Human Evaluation Results:} The \texttt{GPT-4o}, \texttt{Qwen2-7B-Instruct}, and \texttt{O1-preview} models received the highest human evaluation scores of 3.85, suggesting a high degree of alignment between their generated judgments and the legal reasoning in actual case outcomes. Notably, despite achieving only modest scores on automated metrics (with ROUGE-1 scores around 0.13 and BLEU scores of 0.00), these models demonstrate an ability to produce coherent and contextually appropriate responses in legal contexts, as perceived by human evaluators. This underscores the potential of these models to offer valuable insights in complex legal scenarios, even when their textual similarity to reference judgments is limited.

\textbf{Automated Evaluation Results:} Examining the ROUGE and BLEU scores reveals a different dimension of model performance. \texttt{Gemma2-9B}, \texttt{Phi-3.5-mini-instruct}, and \texttt{Mistral-7B-instruct-v0.3} achieved the highest scores on ROUGE-1 (0.39, 0.38, and 0.38 respectively), indicating strong overlap with n-grams in reference texts. However, their BLEU scores remain relatively low (around 0.03 to 0.07), suggesting that while these models generate segments similar to reference texts, they may lack fluency or consistency throughout the entire output. Interestingly, \texttt{lawyer-llama-13b-v2} scored the highest on ROUGE-2 (0.19) and achieved a BLEU score of 0.05, reflecting slightly better cohesion in the generated text.

Among the evaluated models, \texttt{GPT-4o}, \texttt{Qwen2-7B-Instruct}, and \texttt{O1-preview} are distinguished by their high human evaluation scores, suggesting a better understanding of legal case nuances. On the other hand, models like \texttt{Gemma2-9B} and \texttt{lawyer-llama-13b-v2} exhibit superior ROUGE scores, indicating precise lexical overlap but possibly lacking in broader contextual accuracy as judged by human evaluators. These results suggest that while algorithmic scores provide useful benchmarks, human assessments are essential to evaluate the actual applicability of LLMs in the legal domain, where interpretative accuracy is critical.

\subsubsection{Performance on English Legal Texts}The performance of each model on English legal texts was assessed using ROUGE-1, ROUGE-2, ROUGE-L, BLEU, and human evaluation scores. Below, we discuss the findings, emphasizing both algorithmic and human evaluation outcomes.

\begin{table*}[ht]
\centering
\caption{Performance of LLMs on English Legal Texts}
\label{tab:english_performance}
\begin{tabular}{|l|c|c|c|c|c|}
\hline
\textbf{Model} & \textbf{English\_ROUGE-1} & \textbf{English\_ROUGE-2} & \textbf{English\_ROUGE-L} & \textbf{English\_BLEU} & \textbf{English\_Evaluation} \\ \hline
Gemma2-9B & 0.38 & 0.36 & 0.38 & 0.02 & 3.54 \\ \hline
GLM-4-9B-chat & 0.34 & 0.14 & 0.16 & 0.00 & 3.54 \\ \hline
GPT-4o & 0.23 & 0.07 & 0.21 & 0.01 & 3.54 \\ \hline
LawGPT\_zh & 0.17 & 0.05 & 0.09 & 0.00 & 2.15 \\ \hline
lawyer-llama-13b-v2 & 0.42 & 0.38 & 0.42 & 0.05 & 2.23 \\ \hline
llama3.2-3B-instruct & 0.25 & 0.10 & 0.17 & 0.06 & 2.38 \\ \hline
Mistral-7B-instruct-v0.3 & 0.27 & 0.12 & 0.15 & 0.04 & 3.62 \\ \hline
O1-preview & 0.31 & 0.13 & 0.29 & 0.07 & 4.08 \\ \hline
Phi-3.5-mini-instruct & 0.44 & 0.41 & 0.44 & 0.04 & 3.08 \\ \hline
Qwen2-7B-Instruct & 0.31 & 0.13 & 0.14 & 0.00 & 3.85 \\ \hline
\end{tabular}
\end{table*}

\textbf{Human Evaluation Results:} The highest human evaluation score of 4.08 was achieved by \texttt{O1-preview} in the English legal text analysis, with \texttt{Qwen2-7B-Instruct} following closely at 3.85, and several models (\texttt{Gemma2-9B}, \texttt{GLM-4-9B-chat}, and \texttt{GPT-4o}) each scoring 3.54. These high scores indicate that these models are capable of producing judgments that align closely with human reasoning in real case scenarios, especially in English. Notably, \texttt{O1-preview} also achieved the highest human evaluation score for Chinese legal texts, tying with \texttt{GPT-4o} and \texttt{Qwen2-7B-Instruct} at 3.85, which highlights its robustness across both languages. 

However, a comparison reveals that overall, models tended to receive higher human evaluation scores on English texts. For example, \texttt{Gemma2-9B} scored 3.54 on English texts but only 3.00 on Chinese, suggesting that it may perform better in English when assessing legal judgment accuracy. Similarly, \texttt{lawyer-llama-13b-v2} received a noticeably lower score of 2.92 on Chinese texts compared to its English score of 2.23. The high scores of \texttt{O1-preview} across both languages are particularly notable, as they suggest that it consistently generates outputs deemed accurate and contextually appropriate in both Chinese and English legal domains.

\textbf{Automated Evaluation Results:} In terms of algorithmic metrics, \texttt{Phi-3.5-mini-instruct} and \texttt{lawyer-llama-13b-v2} emerged as top performers, with ROUGE-1 scores of 0.44 and 0.42, respectively. Additionally, \texttt{Phi-3.5-mini-instruct} had the highest ROUGE-2 and ROUGE-L scores (0.41 and 0.44), indicating substantial n-gram overlap with reference texts, which suggests a high degree of lexical similarity to the ground truth legal judgments. However, these models’ BLEU scores remain relatively low (0.04 to 0.05), suggesting limitations in generating fluent and coherent sequences across the entire output, particularly for complex legal language.

For English legal texts, \texttt{O1-preview} and \texttt{Qwen2-7B-Instruct} were particularly notable for their high human evaluation scores, highlighting their potential for generating legally relevant and contextually accurate judgments. However, models such as \texttt{Phi-3.5-mini-instruct} and \texttt{lawyer-llama-13b-v2} demonstrated superior ROUGE performance, reflecting strong lexical similarity but a possible lack of comprehensive contextual understanding, as reflected in their lower human scores. 

\subsubsection{Overall Performance}

To provide a comprehensive view of the models' performance across both Chinese and English legal texts, we calculated the overall scores for ROUGE-1, ROUGE-2, ROUGE-L, BLEU, and human evaluation. The results are summarized in Table \ref{tab:overall_performance}.

\begin{table*}[ht]
\centering
\caption{Overall Performance of LLMs}
\label{tab:overall_performance}
\begin{tabular}{|l|c|c|c|c|c|}
\hline
\textbf{Model} & \textbf{Overall\_ROUGE-1} & \textbf{Overall\_ROUGE-2} & \textbf{Overall\_ROUGE-L} & \textbf{Overall\_BLEU} & \textbf{Overall\_Evaluation} \\ \hline
Gemma2-9B & 0.39 & 0.26 & 0.39 & 0.03 & 3.27 \\ \hline
GLM-4-9B-chat & 0.31 & 0.15 & 0.20 & 0.00 & 3.35 \\ \hline
GPT-4o & 0.18 & 0.04 & 0.15 & 0.01 & 3.69 \\ \hline
LawGPT\_zh & 0.22 & 0.07 & 0.12 & 0.02 & 2.00 \\ \hline
lawyer-llama-13b-v2 & 0.37 & 0.28 & 0.37 & 0.05 & 2.58 \\ \hline
llama3.2-3B-instruct & 0.28 & 0.10 & 0.16 & 0.05 & 2.00 \\ \hline
Mistral-7B-instruct-v0.3 & 0.32 & 0.13 & 0.17 & 0.06 & 3.08 \\ \hline
O1-preview & 0.22 & 0.07 & 0.19 & 0.04 & 3.96 \\ \hline
Phi-3.5-mini-instruct & 0.41 & 0.27 & 0.41 & 0.03 & 2.62 \\ \hline
Qwen2-7B-Instruct & 0.29 & 0.15 & 0.19 & 0.00 & 3.85 \\ \hline
\end{tabular}
\end{table*}

Across both languages, the \texttt{O1-preview} model achieved the highest overall human evaluation score of 3.96, demonstrating strong alignment with human judgment across diverse legal cases. This performance suggests that \texttt{O1-preview} is particularly capable of producing contextually appropriate and legally relevant outputs, which makes it a standout model in terms of general applicability.

Automated metrics, such as ROUGE and BLEU, indicate mixed outcomes across models. For instance, \texttt{Phi-3.5-mini-instruct} scored the highest in terms of ROUGE-1 and ROUGE-L (0.41), suggesting its ability to capture relevant content from case data. However, its relatively modest human evaluation score of 2.62 implies that the content generated may lack some critical human-judgment nuances despite high lexical overlap. Similarly, \texttt{lawyer-llama-13b-v2} achieved a strong automated score with ROUGE-2 and ROUGE-L scores of 0.28 and 0.37, respectively, but had a lower human score (2.58), suggesting that while it generates content with high lexical precision, its alignment with human interpretive depth in legal cases might be limited.

In general, models such as \texttt{Gemma2-9B} and \texttt{Qwen2-7B-Instruct} demonstrated moderate performance across both automated and human evaluations, while models like \texttt{LawGPT\_zh} and \texttt{llama3.2-3B-instruct} exhibited lower scores in both areas, indicating potential areas for improvement, particularly in complex legal judgment contexts. The analysis underscores that while automated metrics provide insights into model accuracy, human evaluations remain essential for understanding the practical utility of these models in nuanced legal applications.

\section{Challenges}

\subsection{Data privacy}
Cases in the legal domain often involve individuals’ sensitive information, including personal identity, financial status, and medical records. When using this data for model training, there is a risk that the model may unintentionally expose people' sensitive information during content generation, potentially leading to data leakage. To effectively safeguard data privacy, the design and training processes of the model must prioritize the protection of data. It is essential to ensure that the output results do not disclose personal information. Additionally, the research and development team should implement a rigorous data processing and review mechanism for the model's outputs. This will help minimize risks and ensure compliance and security in the application of LLMs in the legal field.

\subsection{The definition of legal liability}

The delineation of legal liability when utilizing LLMs for legal advice and decision-making remains unclear. Although developers typically emphasize the limitations and potential risks of their models upon release and strive to mitigate legal issues during the training process, unintended consequences can still arise.
When a model provides advice or analysis that leads to undesirable outcomes, the question of liability arises: who should be held accountable? Is it the developer, the user, or the model itself? There is currently no consensus on whether users should be liable for decisions made based on model outputs, highlighting the need for further policy discussions and the establishment of a comprehensive legal framework. Such measures are urgently required to ensure the sustainability and security of LLMs in legal practice.

\subsection{Ethical and moral issues}

Due to the diverse sources of data, these models can introduce biases, which may result in unfair outputs. In the legal field, where fairness and impartiality are crucial, ensuring that models remain neutral during case analysis and preventing potential discrimination and injustice is an urgent concern.
Moreover, the lack of transparency in model-generated results complicates users' ability to assess their reliability. This highlights the need for a robust ethical review mechanism in the legal domain to ensure that model outputs adhere to relevant laws, regulations, and ethical standards. By implementing such a mechanism, we can help ensure that the use of LLMs aligns with the principles of justice and accountability in legal practice.

\subsection{Technical limitations}

Although LLMs have demonstrated impressive capabilities in language processing and information analysis, their application in the legal domain still faces significant technical limitations. For instance, these models can struggle with understanding legal terminology, grasping the context of cases, and analyzing complex legal scenarios, which may lead to errors. Additionally, their lack of interpretability creates uncertainty for legal practitioners who rely on their recommendations. This uncertainty can adversely affect the quality of legal decisions and undermine the reliability of legal practice.
To address these challenges, it is crucial to develop more interpretable models and to incorporate human expertise in the decision-making process. By combining the strengths of these models with the nuanced judgment of legal professionals, we can enhance the accuracy and reliability of legal applications.

\subsection{Legislative differences}

As LLMs are adopted globally, differences in regulatory policies across countries may lead to inconsistencies in legal practice. Some countries may have stricter requirements for data privacy protection, while others may focus more on technological innovation and industrial development. Such policy differences can create compliance risks and inconveniences in the application of LLMs in legal services, challenging their widespread adoption and use.
Therefore, when using LLMs in law-related fields, it is essential to fully consider the application context. It is important to address how to avoid situations where the model produces correct results but has an inaccurate scope of application, or where it complies with local laws and regulations but generates erroneous outputs. This not only affects the effectiveness and reliability of legal services but also directly impacts the fairness of legal practice. Additionally, the ongoing updates to legal systems impose higher requirements on these models, necessitating continuous attention from relevant professionals.

\section{Discussion}
In conclusion, while LLMs show considerable potential in assisting with the understanding and processing of legal texts, their limitations in accurately interpreting complex legal language and reasoning remain clear. These models struggle to fully grasp the subtle nuances of legal concepts and their application in specific cases, indicating a need for improvements in training methodologies—particularly in integrating domain-specific legal knowledge and strengthening reasoning capabilities. To fully realize the potential of LLMs in supporting legal professionals, further research and development are necessary.
\section{Acknowledgements}
\noindent This should be a simple paragraph before the bibliography to thank those individuals and institutions who have supported your work on this article.

\bibliographystyle{IEEEtran.bst}
\bibliography{references}



\end{document}